\pdfoutput=1
\documentclass{article}
\usepackage{rldmsubmit,palatino}
\usepackage{times}  
\usepackage{helvet}  
\usepackage{courier}  
\usepackage{url}  
\usepackage{graphicx}  
\usepackage{titlesec}
\usepackage{mathtools}
\usepackage{comment}
\usepackage[11pt]{moresize}

\usepackage{graphicx}

\makeatletter
\newcommand*\bigcdot{\mathpalette\bigcdot@{.5}}
\newcommand*\bigcdot@[2]{\mathbin{\vcenter{\hbox{\scalebox{#2}{$\m@th#1\bullet$}}}}}
\makeatother

\titlespacing*{\paragraph}{0pt}{1.5ex plus 1ex minus .2ex}{1.3ex plus .2ex}
 
\usepackage{times}
\usepackage{xcolor}
\usepackage{soul}
\usepackage{nicefrac}
\usepackage{mathtools}

\usepackage{graphicx}
\usepackage{subfig}
\usepackage{caption}
\usepackage{amsmath} 
\usepackage{amssymb}
\usepackage{amsthm}
\usepackage{mathabx}
\usepackage{array}
\usepackage{bbm}
\usepackage[ruled,linesnumbered,vlined]{algorithm2e}
\usepackage{calrsfs}
\usepackage[capitalize]{cleveref}

\usepackage[belowskip=-5pt,aboveskip=5pt]{caption}
\usepackage{tkz-graph}
\usepackage{tikz}
\usetikzlibrary{arrows,automata,positioning}
\tikzset{
	LabelStyle/.style = {minimum width = 1em, fill = white!50, opacity=0.8,
		text = black, font = \tiny },
	VertexStyle/.append style = { inner sep=2pt,
		font = \bfseries},
	EdgeStyle/.append style = {->, bend left} }

\crefname{algocf}{Algorithm}{Algorithm}
\Crefname{algocf}{Algorithm}{Algorithm}

\DeclarePairedDelimiter{\norm}{\lVert}{\rVert}

\makeatletter

\newcommand*{\rom}[1]{\expandafter\@slowromancap\romannumeral #1@}
\makeatother
\newcommand{\E}{\mathbb{E}}
\renewcommand{\P}{\mathbb{P}}

\newcommand{\opt}{^\star}

\newcommand{\tr}{^{\mathsf{T}}}

\newcommand{\Real}{\mathbb{R}}
\renewcommand{\ss}{~:~}

\usepackage{natbib}
\setcitestyle{square}

\newcommand{\RNum}[1]{\uppercase\expandafter{\romannumeral #1\relax}}

\newcommand{\states}{\mathcal{S}}
\newcommand{\actions}{\mathcal{A}}
\newcommand{\mdp}{\mathcal{M}}
\newcommand{\data}{\mathcal{D}}
\newcommand{\pset}{\mathcal{P}}

\newcommand{\aset}{\mathcal{P}}

\newcommand{\vset}{\mathcal{V}}

\theoremstyle{plain}

\theoremstyle{definition}

\title{Robust Exploration with Tight Bayesian Plausibility Sets}
\author{
Reazul H. Russel\\
Department of Computer Science\\
University of New Hampshire\\
Durham, NH 03824 \\
\texttt{rrussel@cs.unh.edu} \\
\And
Tianyi Gu\\
Department of Computer Science\\
University of New Hampshire\\
Durham, NH 03824 \\
\texttt{gu@cs.unh.edu} \\
\And
Marek Petrik\\
Department of Computer Science\\
University of New Hampshire\\
Durham, NH 03824 \\
\texttt{mpetrik@cs.unh.edu} \\
}

\date{}
\begin{document}
\maketitle


\begin{abstract}
Optimism about the poorly understood states and actions is the main driving force of exploration for many provably-efficient reinforcement learning algorithms. We propose optimism in the face of sensible value functions (OFVF)- a novel \emph{data-driven} Bayesian algorithm to constructing \emph{Plausibility} sets for MDPs to explore robustly minimizing the worst case exploration cost. The method computes policies with tighter optimistic estimates for exploration by introducing two new ideas. First, it is based on Bayesian posterior distributions rather than distribution-free bounds. Second, OFVF does not construct plausibility sets as simple confidence intervals. Confidence intervals as plausibility sets are a sufficient but not a necessary condition. OFVF uses the structure of the value function to optimize the \emph{location} and \emph{shape} of the plausibility set to guarantee upper bounds directly without necessarily enforcing the requirement for the set to be a confidence interval. OFVF proceeds in an episodic manner, where the duration of the episode is fixed and known. Our algorithm is inherently Bayesian and can leverage prior information. Our theoretical analysis shows the robustness of OFVF, and the empirical results demonstrate its practical promise.
\end{abstract}


\acknowledgements{
This work was supported by the National Science Foundation under Grant No. IIS-1717368 and IIS-1815275.
}

\startmain 

\section{Introduction}

Markov decision processes (MDPs) provide a versatile methodology for modeling dynamic decision problems under uncertainty~\citep{Bertsekas1996,Sutton1998,Puterman2005}. A perfect MDP model for many reinforcement learning problems is not known precisely in general. Instead, a reinforcement learning agent tries to maximize its cumulative payoff by interacting in an unknown environment with an effort to learn the underlying MDP model. It is important for the agent to explore
sub-optimal actions to accelerate the MDP learning task which can help to optimize long-term performance. But it is also important to pick actions with highest known rewards to maximize short-run performance. So the agent always needs to balance between them to boost the performance of a learning algorithm during learning. 

\emph{Optimism in the face of uncertainty (OFU)} is a common principle for most reinforcement learning algorithms encouraging exploration~\citep{Auer2010a,Brafman2001,Kearns1998a}. The idea is to assign a very high exploration bonus to poorly understood states and actions. As the agent visits and gathers statistically significant evidence for these states-actions, the uncertainty and optimism decreases converging to reality. Many RL algorithms including \emph{Explicit Explore or Exploit $(E^3)$}~\citep{Kearns1998a}, \emph{R-{\ssmall MAX}}~\cite{Brafman2001}, \emph{UCRL2}~\citep{Auer2006,Auer2010a}, \emph{MBIE}~\citep{Strehl2008,Strehl2004,Strehl2004a,Wiering1998} build on the idea of optimism guiding the exploration. Probability matching class of algorithms like \emph{Posterior Sampling for reinforcement learning (PSRL)}~\citep{Osband2016, Osband2013, Strens2002} performs exploration with a proportional likelihood to the underlying true parameters. PSRL algorithm is simple, computationally efficient and can utilize any prior structural information to improve exploration. These algorithms provide strong theoretical guarantees with polynomial bound on sample complexity.

During exploration, it is possible for an agent to be overly optimistic about a potentially catastrophic situation and end up there paying an extremely high price (e.g. a self driving car hits a wall, a robot falls off the cliff etc.). Exploring and learning such a situation may not payoff the price. It can be wise for the agent to be robust and avoid those situations minimizing the worst-case exploration cost$-$ which we call robust exploration. OFU and PSRL algorithms are optimistic by definition and cannot guarantee robustness while exploring. The main contribution of this paper is OFVF, an optimistic counter part of RSVF~\citep{Russel2018}. OFVF is a Bayesian approach of constructing plausibility sets for robust exploration.

The paper is organized as follows: \cref{sec:prbolem} formally defines the problem setup and goals of the paper. \cref{sec:IE} reviews some existing methods to construct the plausibility sets and their extension to Bayesian setting. OFVF is proposed and analyzed in \cref{sec:ofvf}. Finally, \cref{sec:experiments} presents empirical performance on several problem domains.


\section{Problem Statement} \label{sec:prbolem}

We consider the problem of learning a finite horizon Markov Decision Process $\mdp$ with states $\states = \{1, \ldots, S \}$ and actions $\actions = \{1, \ldots, A \}$. $p: \states \times \actions \to \Delta^\states$ is a transition function, where $p^a_{ss'}$ is interpreted as the probability of ending in state $s'\in\states$ by taking an action $a\in\actions$ from state $s\in\states$. We omit $s'$ when the next state is not deterministic and denote the transition probability as $p_{sa}\in\Real^S$. $R: \states \times \actions \to \Real$ is a reward function and $R^a_{ss'}$ is the reward for taking action $a\in\actions$ from state $s\in\states$ and reaching state $s'\in\states$. Each MDP $\mdp$ is associated with a discount factor $0\le\gamma\le1$ and a distribution of initial state probabilities $p_0$. We consider an episodic learning process where $L$ is the number of episodes and $H$ is the number of periods in each episode. A policy $\pi = (\pi_0,\ldots,\pi_{H-1})$ is a set of functions mapping a state $s\in\states$ to an action $a\in\actions$. We define a value function for a policy $\pi$ as:
\begin{equation} \label{eq:state_value}
V^\pi_h(s) := \sum_{s'} P^{\pi(s)}_{ss'} [ r_h + V(s') ]
\end{equation}
The optimal value function is defined by $V\opt_h(s) = \max_\pi V^\pi_h(s)$ and the optimal policy is defined by $\pi\opt(s) = \arg\max_{a\in\actions} p^a_{ss'}V(s') \text{, } \forall s'\in\states: p^a_{ss'}>0$.

Optimistic algorithms encouraging exploration find the probability distribution $\tilde{P}_{sa}$ for each state and action within an interval of the empirically derived distribution $\bar{p}_{sa} = \E[\cdot|s,a]$, which defines the plausible set $\pset_{sa}$ of MDPs. They then solve an optimistic version
of \cref{eq:state_value} within $\pset_{sa}$ that leads to the policy with highest reward.
\begin{equation}
V\opt_h(s,a) := \max_{p_{sa} \in \pset_{sa}} \sum_{s'}p^{\pi(s)}_{ss'}[r_h + V\opt (s') ]
\end{equation}
We evaluate the performance of the agent in terms of worst-case \emph{cumulative regret}, which is the maximum total regret incurred by the agent upto time $T$ for a policy $\pi\opt_l$:
\begin{equation}
Regret(T, \pi\opt_l) = \sum_{l=0}^{T/H-1} \sup\bigg[ \sum_{s\in\states} p_0(s)\big( V\opt(s) - V^{\pi\opt_l}(s) \big) \bigg]
\end{equation}

Where $V\opt(s)$ is the true value w.r.t $\mdp^*$.

\section{Interval Estimation for Plausibility Sets} \label{sec:IE}
In this section, we first describe the standard approach to constructing plausibility sets as distribution free confidence intervals. We then propose its extension to Bayesian setting and present a simple algorithm to serve that purpose. It is important to
note that distribution-free bounds are subtly different from the Bayesian bounds, the Bayesian safety guarantee holds conditional on a given dataset $\data$ while the distribution-free hold
across the sets. This makes the guarantees qualitatively different and difficult to compare.

\subsection{Plausibility Sets as Confidence Intervals} \label{ssec:freq_pset}
It is common in the literature to use $L_1$ norm as the distribution-free bound. This bound is constructed around the empirical mean of the transition probability $\bar{p}_{s,a}$ by applying the Hoeffding inequality~\citep{Auer2010a,Petrik2016,Wiesemann2013,Strehl2004}.
\[
\pset_{sa} = \bigg\{ \lVert \tilde{p}_{sa} - \bar{p}_{sa} \rVert_1 \le \sqrt{\frac{2}{n_{s,a}}\log\frac{SA2^S}{\delta}} \bigg\}
\]
where $\bar{p}_{sa}$ is the mean transition computed from D, $n_{s,a}$ is the number of times the agent arrived in state $s'$ after taking action $a$ in state $s$, $\delta$ is the required probability of the interval and $\lVert \bigcdot \rVert_1$ is the $L_1$ norm. An important limitation of this approach is that, the size of $\pset_{sa}$ grows linearly with the number of states, which makes it practically useless in general.

\subsection{Bayesian Plausibility Sets} \label{ssec:bayes_pset}

The Bayesian plausibility sets take the same interval estimation idea and extend it into Bayesian setting, which is analogous to \emph{credible intervals} in Bayesian statistics. Credible intervals are constructed with the posterior probability distributions and they are fixed $-$ not a random variable, given the data $\data$. Instead the estimated transition probabilities maximizing the rewards are random variables. To construct a plausibility set, we optimize for the smallest credible region around the mean transition probability with the assumption that a smaller region will lead to a tighter upper bound estimate.
Formally, the optimization problem to compute $\psi_{s,a}$ for each state s and action a is:
\begin{equation} \label{eq:optimization_bci}
\min_{\psi\in\Real_+} \left\{\psi \ss \P\left[ \norm{\tilde{p}_{s,a} - \bar{p}_{s,a}}_1 > \psi ~|~ \mathcal{D} \right] < \delta \right\}~,
\end{equation}
where nominal point is $\bar{p}_{s,a} = \E_{\tilde{P}}[\tilde{p}_{s,a} ~|~ \mathcal{D}]$. A Bayesian extension of the celebrated \emph{UCRL}~~\citep{Auer2010a} algorithm is \emph{BayesUCRL}, which we consider for comparison. BayesUCRL algorithm uses a hierarchical Bayesian model that can be used to infer the posterior transition probability over $p\opt$. The plausibility set here is a function of the $\frac{1}{t}$-quantile of the posterior samples. We omit the details of BayesUCRL to conserve space.

\section{OFVF: Optimism in the Face of sensible Value Functions} \label{sec:ofvf}

\begin{algorithm}
	\KwIn{Desired confidence level $\delta$ and posterior distribution $\P_{P\opt}[\cdot ~|~\mathcal{D}]$ }
	\KwOut{Policy with a maximized safe return estimate }
	Initialize current policy $\pi_0 \gets \arg\max_{\pi} \rho(\pi,\E_{P\opt}[P\opt~|~\mathcal{D}])$\;
	Initialize current value $v_0 \gets v^{\pi_0}_{\E_{P\opt}[P\opt~|~\mathcal{D}]}$\;
	Initialize value set $\mathcal{V}_0 \gets \{v_0 \}$ \;	
	\label{line:make_p_1} Construct $\aset_0$ optimal for $\mathcal{V}_0$\;
	Initialize counter $k\gets 0$\;
	\While{\cref{eq:condition_safe} is violated with $\mathcal{V}=\{v_k\}$}{
		Include $v_k$ that violates \cref{eq:condition_safe}: $\vset_{k+1} \gets \vset_k \cup \{ v_k \}$ \;
		\label{line:make_p_2} Construct $\aset_{k+1}$ optimized for $\vset_{k+1}$\;
		Compute optimistic value function $v_{k+1}$ and policy $\pi_{k+1}$ for $\aset_{k+1}$\;
		$k \gets k + 1$ \;
	}
	\Return $(\pi_k, p_0\tr v_k)$ \;
	\caption{OFVF} \label{alg:OFVF}
\end{algorithm} 

OFVF uses samples from a posterior distribution, similar to a Bayesian confidence interval, but it relaxes the safety requirement as it is sufficient to guarantee for each state $s$ and action $a$ that:
\begin{equation} \label{eq:condition_safe}
\min_{v\in\mathcal{V}} \P_{P\opt} \left[ \max_{p \in \aset_{s,a}} (p - p_{s,a}\opt)\tr v \le 0  ~\middle|~ \mathcal{D} \right] \ge 1-\frac{\delta}{SA}~,
\end{equation}
with $\mathcal{V} = \{ \hat{v}\opt_{\aset} \}$. To construct the set $\aset$ here, the set $\mathcal{V}$ is not fixed but depends on the optimistic solution, which in turn depends on $\aset$. OFVF starts with a guess of a small set for $\mathcal{V}$ and then grows it, each time with the current value function, until it contains $\hat{v}\opt_{\aset}$ which is always recomputed after constructing the ambiguity set $\aset$.

In lines 4 and 8 of \cref{alg:OFVF}, $\aset_i$ is computed for each state-action $s,a \in \states\times\actions$. Center $\bar{p}$ and set size $\psi_{s,a}$ are computed from \cref{eq:center_point} using set $\mathcal{V}$ \& optimal $g_v$ computed by solving \cref{eq:optimal_hyperplane}.
When the set $\mathcal{V}$ is a singleton, it is easy to compute a form of an optimal plausibility set. 
\begin{equation} \label{eq:optimal_hyperplane}
g = \max \left\{ k ~:~ \P_{P\opt} [k \le v\tr p\opt_{s,a}] \ge 1 - \delta/(SA) \right\}
\end{equation}

For a singleton $\mathcal{V}$, it is sufficient for the plausibility set to be a subset of the hyperplane $\{ p \in \Delta^S ~:~ v\tr p = g\opt \}$ for the estimate to be sufficiently optimistic. When $\mathcal{V}$ is not a singleton, we only consider the setting when it is discrete, finite, and relatively small. We propose to construct a set defined in terms of an $L_1$ ball with the minimum radius such that it is safe for every $v\in\mathcal{V}$. Assuming that $\mathcal{V} = \{v_1, v_2, \ldots, v_k \}$, we solve the following linear program:
\begin{equation} \label{eq:center_point}
\begin{gathered}
\psi_{s,a} = \min_{p\in\Delta^S} \Bigl\{ \max_{i=1,\ldots,k} \norm{q_i - p}_1 ~:~ 
\hspace{0.1cm}  v_i\tr  q_i = g_i\opt, q_i \in \Delta^S, i \in 1,\ldots,k  \Bigr\}
\end{gathered}
\end{equation}

In other words, we construct the set to minimize its radius while still intersecting the hyperplane for each $v$ in $\mathcal{V}$.

\begin{figure*}
	\centering
	\begin{minipage}[c]{.45\columnwidth}
		\centering
		\includegraphics[width=\linewidth]{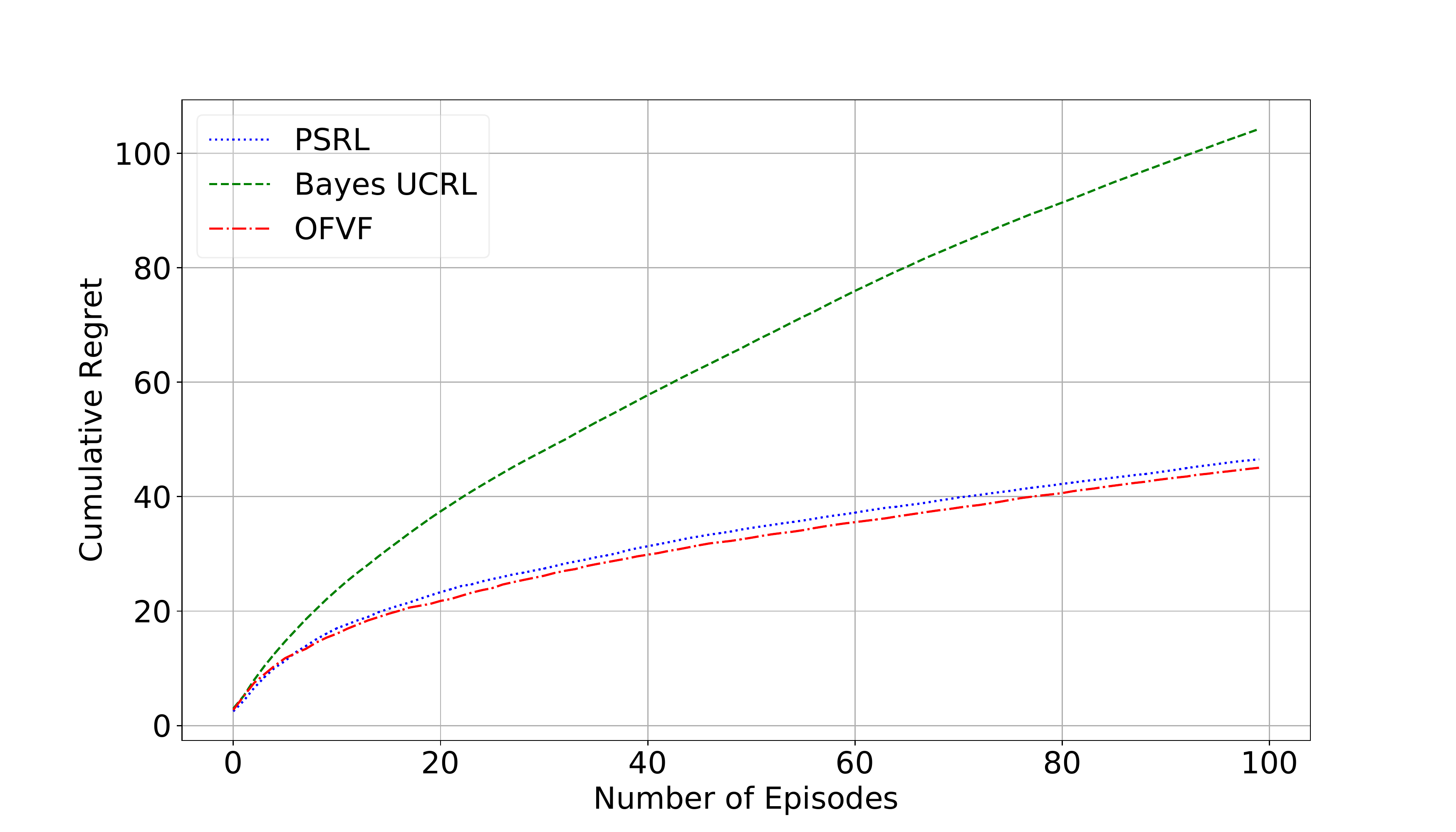}\\
	\end{minipage}%
	\begin{minipage}[c]{.45\columnwidth}
		\centering
		\includegraphics[width=\linewidth]{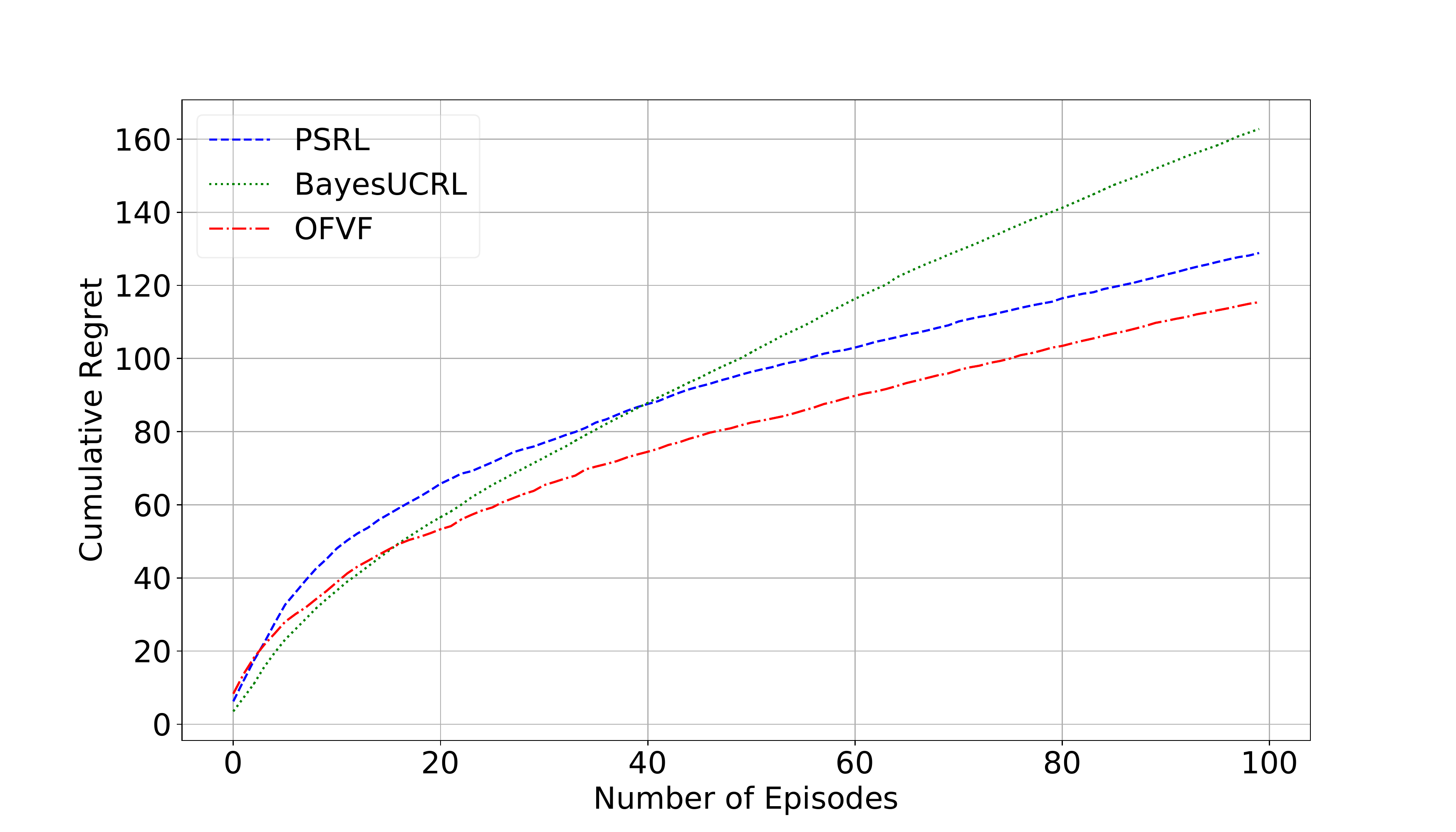}
	\end{minipage}%
	\caption{Cumulative regret for the single-state simple problem. Left) average-case, Right) worst-case.}
	\label{fig:single_state}
\end{figure*}

\section{Empirical Evaluation} \label{sec:experiments}

In this section, we empirically evaluate the estimated returns over episodes. We assume a true model of each problem and generate a number of simulated data sets for the known distribution. We compute the tightest optimistic estimate for the optimal return and compare it with the optimal return for the true model. To judge the performance of the methods, we evaluate both the absolute error of the worst case estimates from optimal, as well the average case estimate from optimal.

We compare our results with BayesUCRL and PSRL algorithms. We omit UCRL from comparison because it performs too poorly compared to other methods. PSRL performs very well in both average and worst case, and as we will see in the experiments, OFVF outperforms BayesUCRL and performs competitively with PSRL. For all the experiments, we use an uninformative Dirichlet prior for the transition probabilities, and run experiments for 100 episodes each containing 100 runs, unless otherwise specified.

\paragraph{Single-state Bellman Update} We initially consider a simple problem with one single non-terminal state. The agent can take three different actions on that state. Each action leads to one of three terminal states with different transition probabilities. The value function for the terminal states are fixed and assumed to be known. \cref{fig:single_state} compares the average-case and worst-case returns computed by different methods. Note that OFVF outperforms all other methods in this simplistic setting. OFVF is able to explore in a robust way maximizing the worst and average case returns.

\begin{figure*}
	\centering
	\begin{minipage}[c]{.45\columnwidth}
		\centering
		\includegraphics[width=\linewidth]{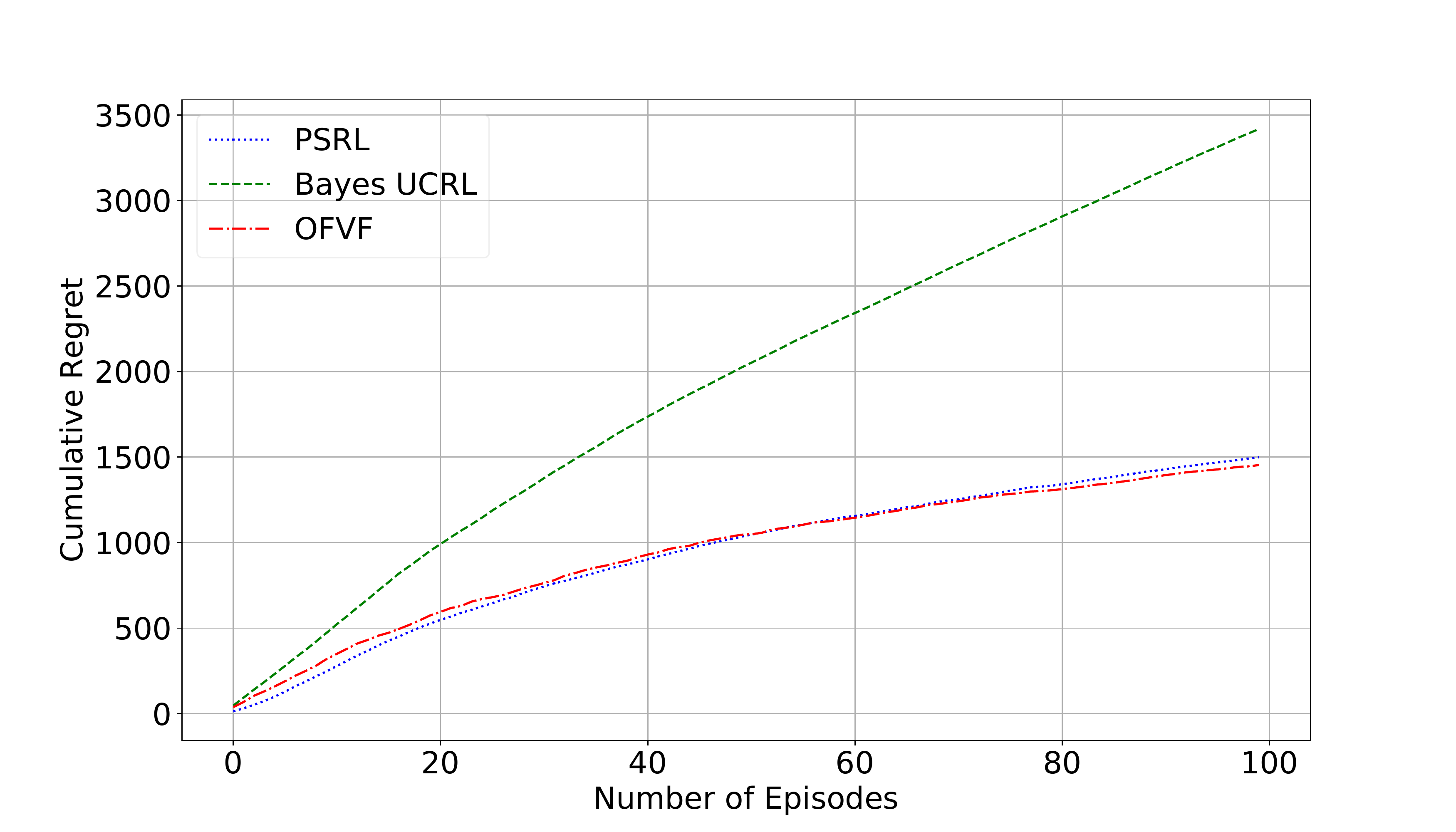}\\
	\end{minipage}%
	\begin{minipage}[c]{.45\columnwidth}
		\centering
		\includegraphics[width=\linewidth]{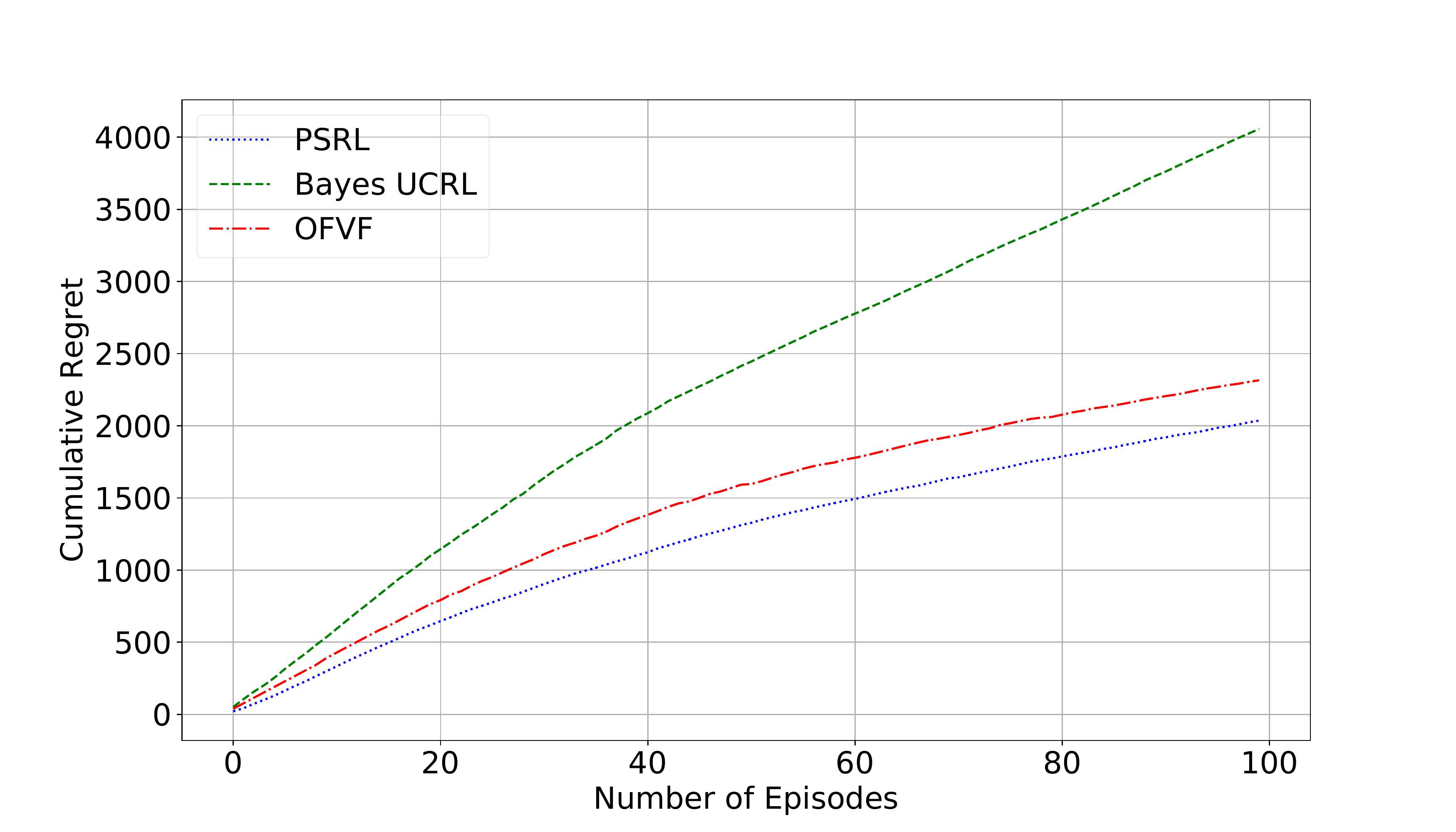}
	\end{minipage}%
	\caption{Cumulative regret for the RiverSwim problem. Left) average-case, Right) worst-case.}
	\label{fig:riverswim}
\end{figure*}

\paragraph{RiverSwim Problem} We compare the performance of different methods in standard example of RiverSwim~\citep{Osband2013, Strehl2004}. The problem is designed requiring hard exploration to find the optimal policy, we omit the full description of the problem to preserve space. \cref{fig:riverswim} compares the average and worst case regrets of different methods. Among optimistic methods, OFVF performs better than BayesUCRL both in average and worst case scenario. But the stochastically optimistic PSRL outperforms all other methods. This is due to the fact that, BayesUCRL and OFVF constructs a plausibility set for each state and action. Even if the plausibility sets are tight, the resulting optimistic MDP is simultaneously optimistic in each state-action, yielding a way too optimistic overall MDP model~\citep{Osband2016}. Thus OFVF can construct tighter plausibility sets for exploration, but still may not match the statistical efficiency of PSRL. This performance however shows that, as an OFU algorithm, OFVF can be reasonably optimistic and can offer competitive performance.


\section{Summary and Conclusion} \label{sec:conclusion}
In this paper, we proposed OFVF, a Bayesian algorithm capable of constructing plausibility sets with better shapes and sizes. Beside the fact that our proposed Bayesian methods are computationally demanding than other distribution free methods, our theoretical and experimental analysis furnished that they can pay-off with much tighter return estimates. We showed that, OFU algorithms can be useful and can be competitive to stochastically optimistic algorithm like PSRL.

\bibliography{reazul_lib}

\end{document}